\documentclass[10pt,twocolumn,letterpaper]{article}

\usepackage[pagenumbers]{wacv} 

\usepackage{url}            
\usepackage{booktabs}       
\usepackage{amsfonts}       
\usepackage{nicefrac}       
\usepackage{microtype}      
\usepackage{xcolor}         
\usepackage{amsmath} 
\usepackage{amssymb}
\usepackage{mathtools}
\usepackage{tabularx}
\usepackage{multirow}
\usepackage{stfloats} 
\usepackage{algorithm}
\usepackage{algpseudocode}
\usepackage{xcolor, colortbl} 
\usepackage{svg}
\usepackage[table]{xcolor}

\usepackage{amsthm}

\usepackage[utf8]{inputenc} 
\usepackage[T1]{fontenc}    

\definecolor{wacvblue}{rgb}{0.21,0.49,0.74}
\usepackage[pagebackref,breaklinks,colorlinks,allcolors=wacvblue]{hyperref}

\def\wacvPaperID{*****} 
\def\confName{WACV}
\def\confYear{2026}

\title{Vision Non-Causal Trapezoidal Mamba: Eliminating Directional Scanning in Vision SSMs with Second-Order Dynamics}
\author{
Anvitha Ramachandran$^{1}$ \qquad Dhruv Parikh$^{1}$ \\ \qquad Haoyang Fan$^{1}$ \qquad Rajgopal Kannan$^{2}$ \qquad Viktor Prasanna$^{1}$ \\
$^{1}$University of Southern California \qquad $^{2}$DEVCOM Army Research Office \\
{\tt\small [alramach, dhruvash, prasanna\}@usc.edu \qquad rajgopal.kannan.civ@army.mil}
}

\begin{document}
\maketitle

\begin{abstract}
\noindent State Space Models (SSMs) have emerged as an alternative to Vision Transformers, yet most vision SSMs inherit directional token scanning from causal sequence modeling. While effective for sequential data, directional scanning introduces spatial bias and orientation-sensitive representations. We present Vision Non-Causal Trapezoidal Mamba (VNCT), a second-order non-causal vision SSM that enables all image tokens to interact in a single pass, eliminating direSctional scanning and achieving low single-image inference latency. VNCT exhibits more orientation-robust representations, showing reduced performance degradation under image rotations and flips, while improving Boundary IoU by up to 3.7 points, leading to more accurate boundary preservation and object localization. Across ImageNet-1K classification, COCO object detection and instance segmentation, and ADE20K semantic segmentation, VNCT consistently outperforms both directional-scanning vision SSMs and first-order non-causal SSMs. These results show that directional scanning is unnecessary for high-performance vision SSMs and that second-order non-causal state-space modeling offers a simple, efficient, and robust alternative for visual recognition. The code for the model will be made available at https://github.com/anvitha305/VNCT.
\end{abstract}

\section{Introduction}

State Space Models (SSMs) have emerged as an efficient alternative to Vision Transformers (ViTs) for modeling long-range dependencies with linear-time complexity. Recent advances, including selective state-space models and State Space Duality (SSD), have significantly improved the expressiveness and efficiency of sequence modeling \cite{gu2023mamba,dao2024transformers}. Building on these developments, vision SSMs such as ViM, VMamba, LocalMamba, and VSSD have demonstrated competitive performance across image classification and dense prediction tasks \cite{zhu2024ViM,liu2024vmamba,huang2024localmamba,VSSD}. These successes have established state-space models as a promising family of vision backbones.

Despite their success, most vision SSMs inherit directional token scanning from causal sequence modeling. Because conventional SSMs operate over one-dimensional sequences, image tokens must be serialized through predefined scan paths before state propagation. While effective, directional scanning imposes an artificial traversal order that introduces spatial bias, orientation-sensitive feature interactions, and often requires multiple scan passes to aggregate bidirectional context. These assumptions are fundamentally misaligned with the non-causal nature of visual scenes, where spatial relationships are independent of token ordering. This raises a fundamental question: \emph{is directional scanning necessary for high-performance vision state-space models, or is it simply an artifact of adapting sequence models to images?}

Recent work has begun addressing this limitation through non-causal state-space formulations. In particular, Vision State Space Duality (VSSD) demonstrated that directional scanning can be replaced with global non-causal aggregation while maintaining competitive visual performance \cite{VSSD}. Meanwhile, Mamba-3 introduced second-order exponential-trapezoidal dynamics that substantially improve sequence modeling capability \cite{lahoti2026mamba}. However, existing non-causal vision SSMs remain limited to first-order dynamics, leaving the potential of second-order non-causal state-space modeling largely unexplored.

We present Vision Non-Causal Trapezoidal (VNCT) Mamba, a hierarchical vision backbone built upon a second-order non-causal state-space formulation. Its core Non-Causal Mamba-3 (NC-M3) layer reformulates the exponential-trapezoidal dynamics of Mamba-3 as a global non-causal aggregation operator, enabling all image tokens to interact in a \emph{single pass} without directional scanning. VNCT further incorporates a low-rank multi-input multi-output (MIMO) parameterization to jointly model spatial and channel-wise interactions. Eliminating directional scanning reduces single-image inference latency while producing more orientation-robust representations.

We evaluate VNCT on ImageNet-1K classification, COCO object detection and instance segmentation, ADE20K semantic segmentation. VNCT outperforms both directional-scanning vision SSMs and first-order non-causal state-space models. Measured single-image inference latency demonstrates that these gains are achieved without sacrificing deployment efficiency. Furthermore, VNCT exhibits substantially stronger robustness to image rotations and flips and improves Boundary IoU by up to 3.7 points, indicating more accurate boundary preservation and object localization. Together, these results demonstrate that directional scanning is unnecessary for high-performance vision SSMs and that second-order non-causal state-space modeling offers a simple, efficient, robust alternative for visual recognition.

Our contributions are summarized as follows:
\begin{itemize}
\item We propose \textbf{NC-M3}, a second-order non-causal state-space layer that eliminates directional scanning and enables single-pass interactions among all image tokens.
\item We develop \textbf{VNCT}, a hierarchical vision backbone that combines second-order non-causal dynamics with a low-rank MIMO parameterization for efficient visual representation learning.
\item We demonstrate that eliminating directional scanning improves single-image inference latency, orientation robustness, and boundary preservation while consistently outperforming existing directional-scanning and first-order non-causal vision SSMs across image classification and dense prediction benchmarks.
\end{itemize}


\section{Related Work}
\label{sec:related_work}

\subsection{Vision State-Space Backbones and Scanning}
State-space models (SSMs) offer a linear-complexity alternative to self-attention, from structured S4 to selective Mamba and Mamba-2/SSD~\cite{gu2022efficient,gu2023mamba,dao2024transformers}, and their vision extensions rival CNNs and Vision Transformers on classification and dense prediction~\cite{zhu2024ViM,liu2024vmamba,huang2024localmamba,pei2024efficientvmamba,plainmamba,shi2024multi}.
All inherit the causal one-dimensional recurrence of sequence models, flattening images along horizontal, vertical, local, or multi-directional scan paths that impose a spatial order not intrinsic to visual data.
Follow-up work---DAMamba, DefMamba, QuadMamba, GroupMamba, SF-Mamba---improves these scans through adaptive, deformable, quadtree, grouped, or token-swapped routing rather than removing them~\cite{li2025damamba,liu2025defmamba,xie2024quadmamba,shaker2025groupmamba,yoshimura2026sfmamba}, confirming that scan order is a central bottleneck yet still treating vision as ordered traversal.
VNCT instead eliminates the traversal: its NC-M3 operator aggregates over all image tokens through a scan-order-independent, non-causal computation.

\subsection{Scan-Free and Non-Causal Vision SSMs}
The closest prior work removes the causal mask altogether. VSSD introduced a non-causal form of State Space Duality for vision, replacing per-token causal hidden states with a shared global state~\cite{VSSD}; while a relevant work, it is built on first-order Mamba-2/SSD dynamics with a SISO-style parameterization, whereas VNCT lifts Mamba-3's second-order exponential-trapezoidal dynamics into the non-causal setting and additionally introduces low-rank MIMO state mixing.
Other scan-reducing designs---Spatial-Mamba's dilated local state fusion, Mamba2D's causal wavefront recurrence, and PRISMamba's concentric-ring averaging, alongside non-SSM mixers such as SEMA and HAMSA---either retain a residual scan or abandon the state-space view~\cite{xiao2025spatialmamba,baty2024mamba2d,hsieh2026partialring,tran2025sema,patro2026hamsa}.
VNCT alone preserves a structured SSM interpretation, parameterizing global aggregation by second-order SSM coefficients rather than convolutional fusion, unweighted averaging, or spectral multiplication.

\subsection{Operator Expressivity and Global Mixing}
VNCT builds on Mamba-3's enrichment of State Space Duality with exponential-trapezoidal discretization, complex-valued updates, and a MIMO formulation~\cite{dao2024transformers,lahoti2026mamba}, whose second-order left/right-endpoint update generalizes the exponential-Euler step of Mamba-1/2 and aligns with evidence that richer transitions improve SSM expressivity~\cite{merrill2024illusion,grazzi2025negeigen}.
Computing global source statistics followed by per-token readout makes the operator algebraically linear-attention-like~\cite{katharopoulos2020transformers,han2024demystify}, but its kernel is fixed by data-dependent SSM coefficients ($\alpha,\beta,\gamma$) rather than free query--key projections; where RALA treats low rank as the bottleneck of linear attention~\cite{fan2025rala}, VNCT resolves it through MIMO cross-channel coupling from within the SSM.
Against hybrids that interleave SSM and attention modules~\cite{hatamizadeh2025mambavision,fu2025segman,lou2025a2mamba} and MambaOut's claim that Mamba is unnecessary for visual recognition~\cite{yu2024mambaout}, VNCT contends that the limitation is the causal directional scan, not state-space modeling itself---so it retains the SSM, strengthens it with second-order non-causal dynamics, and confines self-attention to the final low-resolution stage.

\section{Methodology}
\label{sec:methodology}

\subsection{Overview}
\label{subsec:overview}

Existing vision SSMs process images through directional token scanning, introducing
an artificial traversal order that biases spatial interactions and often requires
multiple scan passes~\cite{liu2024vmamba,VSSD,zhu2024ViM}. We propose \textbf{VNCT}
(\textbf{V}ision \textbf{N}on-\textbf{C}ausal \textbf{T}rapezoidal Mamba), a
hierarchical vision backbone built on a non-causal reformulation of the second-order
exponential-trapezoidal dynamics of Mamba-3~\cite{lahoti2026mamba}. The core
\textit{non-causal lift} transforms the causal Mamba-3 recurrence into a globally
factorized, permutation-invariant aggregation operator, enabling all image tokens to
interact in a single pass without directional scanning. The resulting \textbf{NC-M3}
layer incorporates input-dependent state parameterization, 2D rotary positional
encoding, and an optional low-rank MIMO state-mixing mechanism, and is stacked into
a four-stage hierarchical backbone for classification, detection, instance
segmentation, and semantic segmentation.

\subsection{Preliminaries}
\label{subsec:preliminaries}

\subsubsection{Linear State-Space Models}
\label{subsubsec:ssm}

Let $\mathbf{X} = [\mathbf{x}_1, \ldots, \mathbf{x}_L]^\top \in \mathbb{R}^{L
\times C}$ denote $L$ image tokens with channel dimension $C$, and let
$\mathbf{H} \in \mathbb{R}^{L \times C}$ be the output sequence. A continuous-time
linear SSM evolves a latent state $\mathbf{h}(t)$ as
\begin{equation}
    \dot{\mathbf{h}}(t) = \mathbf{A}\mathbf{h}(t) + \mathbf{b}\,x(t),\qquad
    y(t) = \mathbf{E}\mathbf{h}(t) + D\,x(t).
    \label{eq:ssm_cont}
\end{equation}
Zero-order-hold (ZOH) discretization with step $\Delta$ yields
\begin{equation}
    \mathbf{h}_t = \underbrace{\exp(\Delta\mathbf{A})}_{\boldsymbol{\alpha}_t}
    \mathbf{h}_{t-1}
    + \underbrace{(\Delta\mathbf{A})^{-1}(\exp(\Delta\mathbf{A})-\mathbf{I})
    \Delta\mathbf{b}}_{\boldsymbol{\beta}_t} x_t.
    \label{eq:ssm_discrete_first}
\end{equation}
Efficient architectures such as S4~\cite{gu2022efficient}, Mamba~\cite{gu2023mamba}, and
Mamba-2~\cite{dao2024transformers} restrict $\mathbf{A}$ to diagonal form. The
sequential structure of Eq.~\eqref{eq:ssm_discrete_first} encodes only the prefix
$\{x_1,\ldots,x_t\}$ at each position, introducing scan anisotropy when applied to
2D visual inputs.

\subsubsection{Mamba-3 and Second-Order State Dynamics}
\label{subsubsec:mamba3}

Mamba-3~\cite{lahoti2026mamba} replaces ZOH with trapezoidal quadrature, averaging over both
integration endpoints:
\begin{equation}
    \mathbf{h}_t = \boldsymbol{\alpha}_t\,\mathbf{h}_{t-1}
    + \boldsymbol{\beta}_t\,x_{t-1}
    + \boldsymbol{\gamma}_t\,x_t,
    \label{eq:mamba3}
\end{equation}
where $\boldsymbol{\beta}_t$ and $\boldsymbol{\gamma}_t$ weight the left- and
right-endpoint observations respectively. This second-order correction induces a
richer implicit convolutional kernel and improves sequence modeling
quality~\cite{lahoti2026mamba}, but the recurrence remains causal, enforcing a scan trajectory
on visual inputs. VNCT derives an equivalent order-independent formulation of
Eq.~\eqref{eq:mamba3}'s information-mixing behavior, described next.

\subsection{Non-Causal Lift of Trapezoidal State Dynamics}
\label{subsec:noncausal_reformulation}

The goal is to construct a permutation-invariant operator
\begin{equation}
    \mathbf{H} = \mathcal{F}_{\mathrm{NC}}(\mathbf{X}),\qquad
    \mathbf{H} \in \mathbb{R}^{L \times C},
    \label{eq:operator_goal}
\end{equation}
that captures the information-mixing behavior of Eq.~\eqref{eq:mamba3} while
permitting fully parallel, order-independent evaluation. The \textit{non-causal lift}
preserves the roles of $\boldsymbol{\alpha}$, $\boldsymbol{\beta}$,
$\boldsymbol{\gamma}$ as interaction coefficients while discarding sequential state
dependence and scan ordering.

\subsubsection{Kernel Formulation of the Non-Causal Interaction}
\label{subsubsec:kernel}

We replace the lower-triangular causal transition matrix with a fully dense,
normalized interaction kernel. For output location $i$, source location $j$, and
channel $c$:
\begin{equation}
    \mathcal{W}_{ij,c}
    = \frac{
        \phi\!\left(\boldsymbol{\alpha}_{i,c}\right)^\top
        \psi\!\left(\boldsymbol{\beta}_{j,c} + \boldsymbol{\gamma}_{j,c}\right)
      }{
        \sum_{k=1}^{L}
        \phi\!\left(\boldsymbol{\alpha}_{i,c}\right)^\top
        \psi\!\left(\boldsymbol{\beta}_{k,c} + \boldsymbol{\gamma}_{k,c}\right)
      },
    \label{eq:kernel}
\end{equation}
where $\phi,\psi : \mathbb{R}^D \to \mathbb{R}^D_{\geq 0}$ are positive-valued
feature maps projecting SSM parameters into a $D$-dimensional interaction space.
Here $\phi(\boldsymbol{\alpha}_{i,c})$ acts as a query controlling context
aggregation range --- large $\boldsymbol{\alpha}$ (slow decay) produces large-norm
queries while small $\boldsymbol{\alpha}$ (rapid decay) yields near-local outputs
--- preserving the semantic role of $\boldsymbol{\alpha}$ as a memory-length
controller. The source term $\boldsymbol{\beta}_{j,c}+\boldsymbol{\gamma}_{j,c}$
combines both trapezoidal endpoints and depends only on index $j$, enabling
precomputation independently of $i$ for the linear-time factorization below. The
denominator ensures $\sum_j \mathcal{W}_{ij,c}=1$.

Factoring $\phi(\boldsymbol{\alpha}_{i,c})$ out of the source sums gives the
aggregation rule
\begin{equation}
\begin{aligned}
    h_{i,c}
    &= \sum_{j=1}^{L} \mathcal{W}_{ij,c}\,\tilde{x}_{j,c} \\
    &= \frac{
        \phi\!\left(\boldsymbol{\alpha}_{i,c}\right)^\top\!
        \sum_{j=1}^{L}
        \psi\!\left(\boldsymbol{\beta}_{j,c} + \boldsymbol{\gamma}_{j,c}\right)
        \tilde{x}_{j,c}
      }{
        \phi\!\left(\boldsymbol{\alpha}_{i,c}\right)^\top
        \sum_{k=1}^{L}
        \psi\!\left(\boldsymbol{\beta}_{k,c} + \boldsymbol{\gamma}_{k,c}\right)
      },
\end{aligned}
    \label{eq:aggregation_expanded}
\end{equation}
where $\tilde{x}_{j,c} = b_{j,c}\,x_{j,c}$. Although algebraically similar to
linear attention~\cite{katharopoulos2020transformers,choromanski2020rethinking},
$\phi$ and $\psi$ here operate on SSM discretization parameters rather than arbitrary
input features, inheriting the inductive biases of the Mamba-3 parameterization.

\subsubsection{Factorized Global Aggregation}
\label{subsubsec:factorization}

The inner sums in Eq.~\eqref{eq:aggregation_expanded} are independent of $i$,
enabling linear-time computation via global sufficient statistics
\begin{align}
    \mathbf{S}_c
    &= \sum_{j=1}^{L}
       \psi\!\left(\boldsymbol{\beta}_{j,c} + \boldsymbol{\gamma}_{j,c}\right)
       \tilde{x}_{j,c},
    \label{eq:S}
    \\[4pt]
    \mathbf{Z}_c
    &= \sum_{j=1}^{L}
       \psi\!\left(\boldsymbol{\beta}_{j,c} + \boldsymbol{\gamma}_{j,c}\right),
    \label{eq:Z}
\end{align}
with $\mathbf{S}_c, \mathbf{Z}_c \in \mathbb{R}^D$ summarizing the entire token set
independently of any output index. The aggregation reduces to
\begin{equation}
    h_{i,c}
    = \frac{
        \phi\!\left(\boldsymbol{\alpha}_{i,c}\right)^\top \mathbf{S}_c
      }{
        \phi\!\left(\boldsymbol{\alpha}_{i,c}\right)^\top \mathbf{Z}_c
      }.
    \label{eq:nc_aggregation}
\end{equation}
Computation decomposes into two $\mathcal{O}(LD)$ passes: a source-side pass
accumulating $(\mathbf{S}_c, \mathbf{Z}_c)$ and a target-side pass evaluating
$h_{i,c}$ for all $i$ simultaneously, without ever constructing the $L\times L$
interaction matrix.

\subsection{NC-M3: Non-Causal Mamba-3 Block}
\label{subsec:ncm3_layer}

\begin{figure}[t]
    \centering
    \includegraphics[width=\columnwidth]{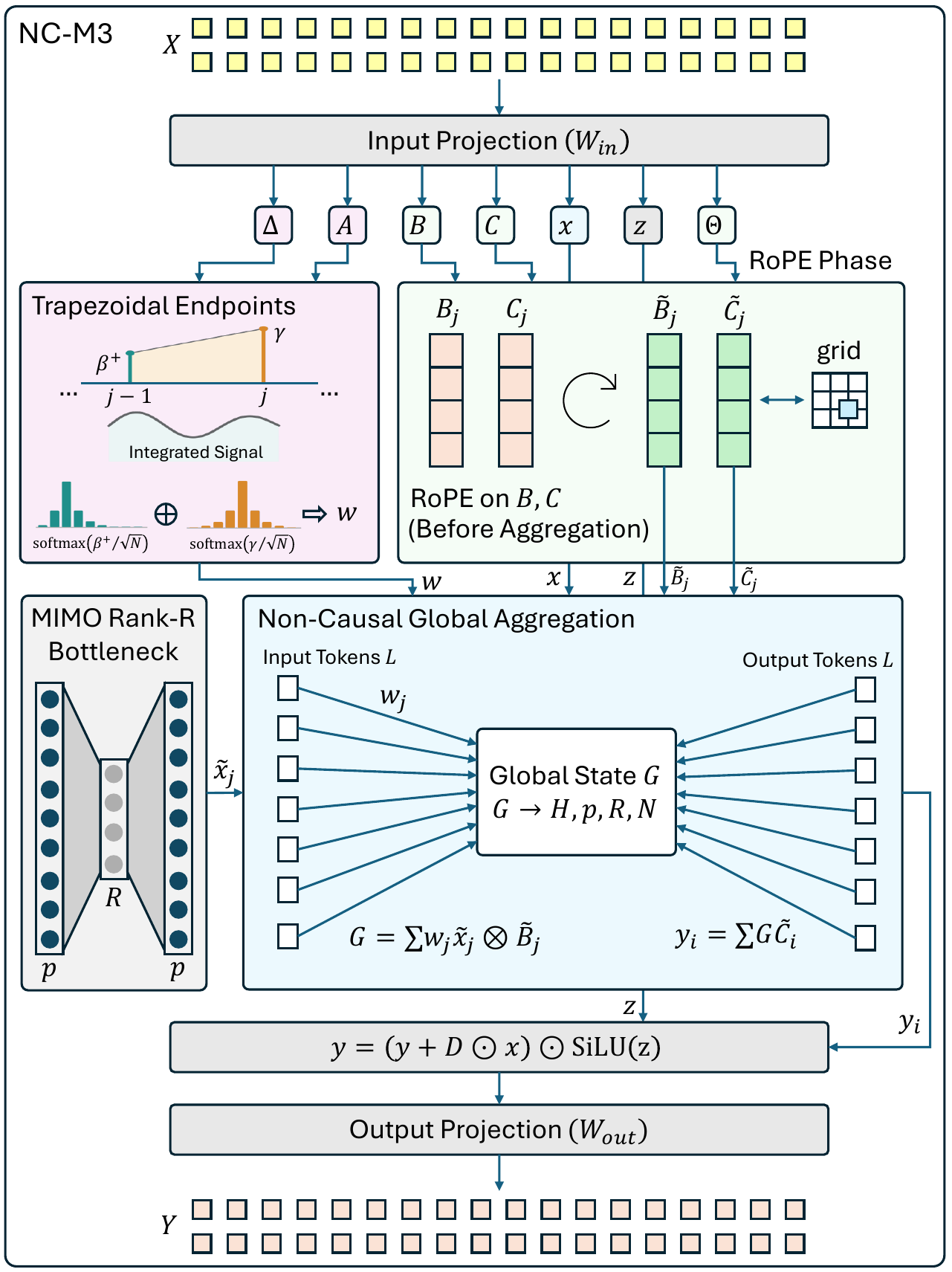}
    \caption{\textbf{The NC-M3 block.} Input tokens are projected into value,
    key/query, gating, and second-order state-parameter streams. Trapezoidal
    endpoints form the fused source weight $w$ and 2D RoPE injects geometry into
    $\mathbf{B},\mathbf{C}$; all tokens then aggregate into a single global state
    $\mathbf{G}$ that every output token reads back, yielding scan-free,
    linear-time global mixing.}
    \label{fig:ncm3}
\end{figure}

The NC-M3 block (Fig.~\ref{fig:ncm3}) instantiates
Eqs.~\eqref{eq:S}--\eqref{eq:nc_aggregation} within a
full feature transformation pipeline. Given input tokens $\mathbf{X} \in
\mathbb{R}^{L \times C}$, NC-M3 produces output tokens $\mathbf{Y} \in
\mathbb{R}^{L \times C}$, following the structural conventions of Mamba and
Mamba-2~\cite{gu2023mamba,dao2024transformers}: expanded inner dimension, parallel
gating and value streams, input-dependent state parameters, and skip-connection
output.

%
%
%
%
%
%
%
%
%
%

\subsubsection{Input Projection and State Parameter Generation}
\label{subsubsec:input_proj}

A single linear projection expands the $C$-dimensional input into all required
streams:
\begin{equation}
    \bigl[\,\mathbf{z},\;\mathbf{x},\;\mathbf{B},\;\mathbf{C},\;\Delta,\;
    \Lambda,\;\boldsymbol{\Theta}\,\bigr]
    = \mathbf{X}\mathbf{W}_{\mathrm{in}}^\top + \mathbf{b}_{\mathrm{in}},
    \label{eq:input_proj}
\end{equation}
with total projected dimension $d_{\mathrm{in}} = 2d_{\mathrm{inner}} + 2NR + 3H
+ S$, where $d_{\mathrm{inner}} = 2C$, $N$ is the SSM state dimension, $H$ the
number of SSM heads, $R$ the MIMO rank, and $S$ the number of RoPE angle dimensions.
Here $\mathbf{z}$ is the gate stream, $\mathbf{x}$ the value stream, $\mathbf{B}$
the source projection ($\psi$ in Eq.~\eqref{eq:kernel}), $\mathbf{C}$ the query
projection ($\phi$ in Eq.~\eqref{eq:kernel}), $\Delta$ the data-dependent step-size
logit, $\Lambda$ the trapezoidal interpolation logit, and $\boldsymbol{\Theta}$ the
RoPE angle stream.

From the projected streams, the three SSM parameters of Eq.~\eqref{eq:mamba3} are
computed as data-dependent, per-head quantities:
\begin{equation}
    \mathrm{DT}_{i,h} = \mathrm{softplus}\!\left(\Delta_{i,h} + b_{\Delta,h}
    \right),\qquad \mathrm{DT} \in \mathbb{R}^{L \times H},
    \label{eq:dt}
\end{equation}
with $A_h = -\mathrm{softplus}(a_h)$ constraining the learned head-wise decay
negative. The three SSM parameters are then
\begin{align}
    \boldsymbol{\alpha}_{i,h} &= \exp\!\bigl(A_h \cdot \mathrm{DT}_{i,h}\bigr),
    \label{eq:alpha}
    \\[2pt]
    \boldsymbol{\gamma}_{i,h} &= \sigma\!\left(\Lambda_{i,h}\right) \cdot
    \mathrm{DT}_{i,h},
    \label{eq:gamma}
    \\[2pt]
    \boldsymbol{\beta}_{i,h} &= \bigl(1 - \sigma\!\left(\Lambda_{i,h}\right)
    \bigr) \cdot \mathrm{DT}_{i,h} \cdot \boldsymbol{\alpha}_{i,h},
    \label{eq:beta}
\end{align}
where $\sigma(\cdot)$ is the sigmoid. Since $A_h<0$ and $\mathrm{DT}_{i,h}>0$,
$\boldsymbol{\alpha}_{i,h}\in(0,1)$. The sigmoid-gated logit partitions
$\mathrm{DT}_{i,h}$ into complementary right-endpoint weight $\boldsymbol{\gamma}$
(fraction $\sigma(\Lambda)$) and left-endpoint weight $\boldsymbol{\beta}$
(fraction $1-\sigma(\Lambda)$, scaled by $\boldsymbol{\alpha}$), recovering the
Mamba-3 trapezoidal discretization.

\subsubsection{Direction-Agnostic Global Aggregation}
\label{subsubsec:global_aggregation}

Since Eq.~\eqref{eq:nc_aggregation} is permutation invariant, spatial information is
injected via 2D rotary positional encoding (RoPE)~\cite{rope} applied to the source
and query projections $\mathbf{B}$ and $\mathbf{C}$, encoding relative spatial
displacement while preserving the linear-time factorization (see
Supplement~\ref{sec:supp_rope}). The fused source weight realizing
$\psi(\boldsymbol{\beta}_{j,c}+\boldsymbol{\gamma}_{j,c})$ is
\begin{equation}
    w_{j,h}
    =
    \mathrm{softmax}_L\!\left(\boldsymbol{\gamma}_{j,h}/\sqrt{N}\right)
    +
    \mathrm{softmax}_L\!\left(\boldsymbol{\beta}^{+}_{j,h}/\sqrt{N}\right),
    \label{eq:w_fused}
\end{equation}
where $\boldsymbol{\beta}^{+}_{j,h}=\boldsymbol{\beta}_{j+1,h}$ is the one-step-ahead
roll of $\boldsymbol{\beta}$. The two softmax terms correspond to the trapezoidal
integration endpoints, yielding a constant normalization term in
Eq.~\eqref{eq:nc_aggregation}.

NC-M3 optionally employs a low-rank MIMO formulation for cross-head interaction at
$\mathcal{O}(CR)$ cost. The value stream is expanded as
\begin{equation}
    \tilde{x}_{i,h,p,r}
    =
    \sum_{p'}x_{i,h,p'}U_{h,r,p'},
    \label{eq:mimo_expand}
\end{equation}
followed by global accumulation and readout:
\begin{align}
    \mathbf{G}_{h,p,r,n}
    &= \sum_{j=1}^{L}
    w_{j,h}\tilde{x}_{j,h,p,r}\tilde{B}_{j,r,n},
    \label{eq:G}
    \\
    y_{i,h,p}
    &= \sum_{r,n}
    \mathbf{G}_{h,p,r,n}\tilde{C}_{i,r,n}.
    \label{eq:readout}
\end{align}
The global state tensor $\mathbf{G}$ is computed once and reused for all tokens,
preserving $\mathcal{O}(LCR)$ overall complexity. MIMO is evaluated in the ablation
studies (Section~\ref{sec:experiments}).

\subsubsection{Output Projection and Residual Update}
\label{subsubsec:output_proj}

Following the readout of Eq.~\eqref{eq:readout}, a per-head skip connection and
gated non-linearity are applied:
\begin{equation}
    \mathbf{y}
    = \bigl(\mathbf{h} + \mathbf{D}\odot\mathbf{x}\bigr)\odot\mathrm{SiLU}(\mathbf{z}),
    \label{eq:gate}
\end{equation}
where $\mathbf{h}$ is the readout, $\mathbf{D}\in\mathbb{R}^H$ is a per-head
learnable skip coefficient, $\mathbf{x}$ the value stream, and $\mathbf{z}$ the gate
stream. After gating, $\mathbf{y}$ is reshaped and projected back to the input
dimension:
\begin{equation}
    \mathbf{Y} = \mathbf{y}\,\mathbf{W}_{\mathrm{out}}^\top,\qquad
    \mathbf{W}_{\mathrm{out}}\in\mathbb{R}^{C\times d_{\mathrm{inner}}},
    \label{eq:output_proj}
\end{equation}
with $\mathbf{Y}\in\mathbb{R}^{L\times C}$ returned as the SSM branch contribution
in Eq.~\eqref{eq:block_ssm}.

\subsection{VNCT Backbone Architecture}
\label{subsec:backbone}

\subsubsection{Hierarchical Architecture}
\label{subsubsec:hierarchy}

\begin{figure*}[t]
    \centering
    \includegraphics[width=\textwidth]{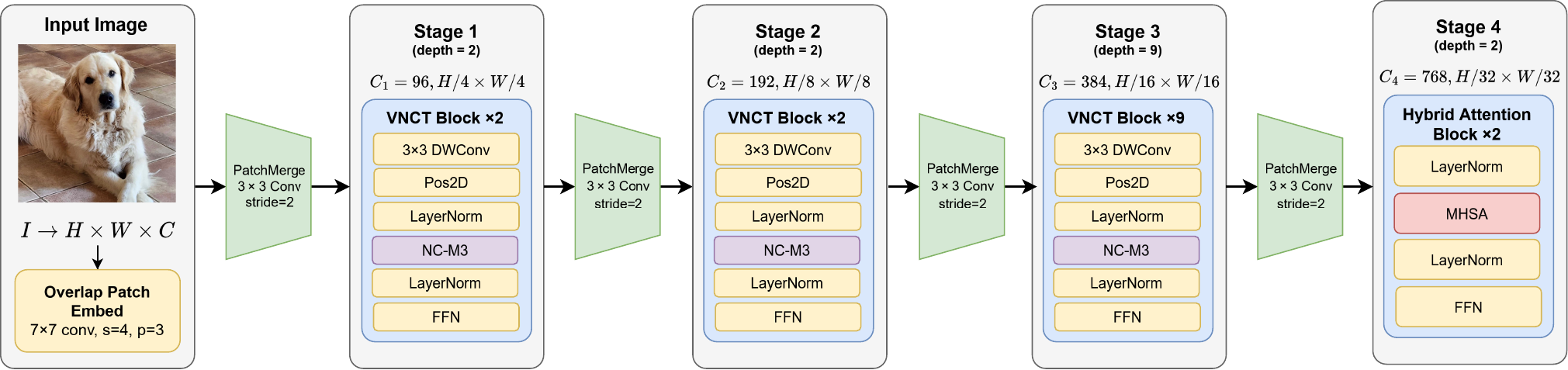}
    \caption{\textbf{Overview of the VNCT backbone.} The input image is processed
    through a four-stage hierarchical architecture with progressive spatial
    downsampling and increasing channel dimensions. Stages~1--3 employ VNCT blocks
    based on the proposed NC-M3 layer, while Stage~4 uses Hybrid-Attention.
    Multi-scale features are produced for downstream dense prediction tasks, and the
    final stage is globally pooled for image classification.}
    \label{fig:vnct_backbone}
    \vspace{-2mm}
\end{figure*}

VNCT adopts a four-stage hierarchical architecture (Fig.~\ref{fig:vnct_backbone})
with progressively decreasing spatial resolution and increasing channel width. An
overlapping patch embedding ($7\times7$ conv, stride~4, padding~3, LayerNorm) yields
\begin{equation}
\mathbf{X}_1\in\mathbb{R}^{\frac{H_{\mathrm{in}}}{4}\times
\frac{W_{\mathrm{in}}}{4}\times C_1},
\label{eq:patch_embed}
\end{equation}
and each subsequent stage applies a $3\times3$ strided convolution to halve spatial
resolution and double the channel dimension. Detailed configurations for all VNCT
variants are provided in the supplementary material.

\subsubsection{VNCT Block}
\label{subsubsec:block}

Each VNCT block (Stages~1--3) applies four components in sequence. First, a Local
Perception Unit (LPU) injects local spatial inductive bias:
\begin{equation}
\mathbf{x}'=\mathbf{x}+\mathrm{LN}\!\left(\mathrm{DWConv}_{3\times3}(\mathbf{x})\right).
\label{eq:lpu}
\end{equation}
2D sinusoidal positional encoding is then added:
\begin{equation}
\mathbf{x}''=\mathbf{x}'+\mathrm{Pos2D}(\mathbf{x}',H,W).
\label{eq:pos2d}
\end{equation}
Global interaction is performed by NC-M3 with a layer-scaled residual:
\begin{equation}
\mathbf{x}'''
=\mathbf{x}''+\boldsymbol{\lambda}_1\odot
\mathrm{NC\text{-}M3}\!\left(\mathrm{LN}(\mathbf{x}''),H,W\right),
\label{eq:block_ssm}
\end{equation}
where $\boldsymbol{\lambda}_1$ is initialized to $10^{-5}$~\cite{cait}. A two-layer
FFN with layer-scaled residual and stochastic depth~\cite{huang2016deep} completes
the block:
\begin{equation}
\mathbf{x}_{\mathrm{out}}
=\mathbf{x}'''+\boldsymbol{\lambda}_2\odot
\mathrm{FFN}\!\left(\mathrm{LN}(\mathbf{x}''')\right).
\label{eq:block_ffn}
\end{equation}

\subsubsection{Multi-Scale Representation}
\label{subsubsec:stage4}

Stage~4 replaces NC-M3 with multi-head self-attention (MHSA), restricting quadratic
attention to the lowest-resolution feature map where its cost is negligible. The four
stages yield a feature pyramid at strides $\{4,8,16,32\}$, making VNCT directly
compatible with UPerNet~\cite{upernet}, FPN~\cite{fpn}, Mask R-CNN~\cite{maskrcnn},
and standard classification heads.
\section{Experiments}
\label{sec:experiments}

\subsection{Implementation Details}

Unless otherwise specified, all experiments are implemented in PyTorch and trained on
8 NVIDIA RTX 6000 Ada GPUs using Distributed Data Parallel with automatic mixed
precision. ImageNet-1K models are trained using a standard AdamW-based recipe with
common data augmentation strategies~\cite{loshchilov2019decoupled,touvron2021training,
convnext,cubuk2020randaugment,zhang2018mixup,yun2019cutmix,zhong2020random,
szegedy2016rethinking}. For downstream detection, instance segmentation, and semantic
segmentation, we follow the official VSSD training protocols~\cite{VSSD} using
Mask R-CNN~\cite{maskrcnn} in MMDetection and UPerNet~\cite{upernet} in
MMSegmentation, initializing all backbones from ImageNet-1K pretrained checkpoints.
Complete hyperparameters are provided in the supplementary material.

\subsection{Image Classification on ImageNet-1K}
\label{subsec:imagenet_exp}

\noindent\textbf{Experimental Setup.}
We evaluate VNCT on ImageNet-1K~\cite{deng2009imagenet} using standard Top-1 accuracy
across micro, tiny, small, and base model scales, comparing against representative
CNNs, Vision Transformers, directional-scanning SSMs, and recent non-causal
state-space models. We additionally report single-image FP16 inference latency on the
same hardware to assess practical deployment efficiency
(Table~\ref{tab:latency}).

\noindent\textbf{Results and Discussion.}
VNCT consistently achieves the strongest classification performance at all evaluated
scales (Tables~\ref{tab:imagenet_micro_tiny},~\ref{tab:imagenet_small_base}),
improving upon both scan-based baselines (VMamba, LocalVMamba) and the first-order
non-causal baseline (VSSD). These results demonstrate that directional sequential
scanning is unnecessary for competitive visual representations, and that the proposed
second-order dynamics provide additional representational capacity over first-order
formulations. Accuracy gains are achieved without sacrificing efficiency: VNCT attains
lower latency than VSSD at the Micro, Tiny, and Small scales and remains competitive
at Base scale (Table~\ref{tab:latency}).

\begin{table}[t]
\centering
\caption{\textbf{ImageNet-1K classification (Micro and Tiny models).}
Top-1 accuracy (\%) on the ImageNet-1K validation set. Method types: Conv
(convolutional), Attn (attention-based), SSM (scan-based state-space), SSD
(first-order non-causal state-space), NCT (second-order non-causal, ours).
Parameter counts and FLOPs are for the classification backbone only.
VNCT achieves the best Top-1 accuracy at both scales.}
\label{tab:imagenet_micro_tiny}
\footnotesize
\setlength{\tabcolsep}{2.5pt}
\renewcommand{\arraystretch}{0.9}
\begin{tabularx}{\columnwidth}{X|c|cc|c}
\toprule
\textbf{Method} & \textbf{Type} & \textbf{\#Param.} & \textbf{FLOPs} & \textbf{Top-1} \\
\midrule
\multicolumn{5}{c}{\textbf{Micro Models}} \\
RegNetY-1.6G~\cite{radosavovic2020designing} & Conv & 11M & 1.6G & 78.0 \\
EffNet-B3~\cite{efficientnet}               & Conv & 12M & 1.8G & 81.6 \\
PVTv2-b1~\cite{pvtv2}                       & Attn & 13M & 2.1G & 78.7 \\
BiFormer~\cite{biformer}                    & Attn & 13M & 2.2G & 81.4 \\
NAT-M~\cite{nat}                            & Attn & 20M & 2.7G & 81.8 \\
SMT-T~\cite{lin2023scale}                   & Attn & 12M & 2.4G & 82.2 \\
Vim-T~\cite{zhu2024ViM}                     & SSM  &  7M & 1.5G & 76.1 \\
LVim-T~\cite{huang2024localmamba}           & SSM  &  8M & 1.5G & 76.2 \\
VSSD-M~\cite{VSSD}                          & SSD  & 14M & 2.3G & 82.5 \\
\rowcolor{gray!15} VNCT-Micro               & NCT  & 15M & 2.5G & \textbf{83.3} \\
\midrule
\multicolumn{5}{c}{\textbf{Tiny Models}} \\
ConvNeXt-T~\cite{convnext}                  & Conv & 29M & 4.5G & 82.1 \\
MambaOut-T~\cite{yu2024mambaout}            & Conv & 27M & 4.5G & 82.7 \\
EffNet-B4~\cite{efficientnet}               & Conv & 19M & 4.2G & 82.9 \\
Swin-T~\cite{Swin}                          & Attn & 29M & 4.5G & 81.3 \\
PVTv2-B2~\cite{pvtv2}                       & Attn & 25M & 4.0G & 82.0 \\
NAT-T~\cite{nat}                            & Attn & 28M & 4.3G & 83.2 \\
VMambaV9-T~\cite{liu2024vmamba}             & SSM  & 31M & 4.9G & 82.5 \\
MSVMamba-T~\cite{shi2024multi}              & SSM  & 33M & 4.6G & 82.8 \\
VSSD-T~\cite{VSSD}                          & SSD  & 24M & 4.5G & 83.7 \\
\rowcolor{gray!15} VNCT-Tiny                & NCT  & 28M & 5.0G & \textbf{84.2} \\
\bottomrule
\end{tabularx}
\vspace{-2mm}
\end{table}

\begin{table}[t]
\centering
\caption{\textbf{ImageNet-1K classification (Small and Base models).}
Top-1 accuracy (\%) at larger model scales. VNCT-Small achieves state-of-the-art
accuracy with fewer FLOPs than VMamba-S; VNCT-Base surpasses VSSD-B while using
fewer parameters and FLOPs. See Table~\ref{tab:imagenet_micro_tiny} for type
abbreviations.}
\label{tab:imagenet_small_base}
\footnotesize
\setlength{\tabcolsep}{2.5pt}
\renewcommand{\arraystretch}{0.9}
\begin{tabularx}{\columnwidth}{X|c|cc|c}
\toprule
\textbf{Method} & \textbf{Type} & \textbf{\#Param.} & \textbf{FLOPs} & \textbf{Top-1} \\
\midrule
\multicolumn{5}{c}{\textbf{Small Models}} \\
ConvNeXt-S~\cite{convnext}   & Conv & 50M & 8.7G  & 83.1 \\
MambaOut-S~\cite{yu2024mambaout} & Conv & 48M & 9.0G & 84.1 \\
Swin-S~\cite{Swin}           & Attn & 50M & 8.7G  & 83.0 \\
VMamba-S~\cite{liu2024vmamba}& SSM  & 44M & 11.2G & 83.5 \\
VSSD-S~\cite{VSSD}           & SSD  & 40M & 7.4G  & 84.5 \\
\rowcolor{gray!15} VNCT-Small & NCT & 44M & 8.0G  & \textbf{84.7} \\
\midrule
\multicolumn{5}{c}{\textbf{Base Models}} \\
ConvNeXt-B~\cite{convnext}   & Conv & 89M & 15.4G & 83.8 \\
MambaOut-B~\cite{yu2024mambaout} & Conv & 85M & 15.8G & 84.2 \\
Swin-B~\cite{Swin}           & Attn & 88M & 15.4G & 83.5 \\
VMambaV9-B~\cite{liu2024vmamba} & SSM & 89M & 15.4G & 83.9 \\
VSSD-B~\cite{VSSD}           & SSD  & 89M & 16.1G & 85.4 \\
\rowcolor{gray!15} VNCT-Base  & NCT  & 86M & 15.5G & \textbf{85.6} \\
\bottomrule
\end{tabularx}
\vspace{-2mm}
\end{table}

\begin{table}[t]
\centering
\caption{\textbf{Single-image FP16 inference latency (ms/image).}
Measured on a single NVIDIA RTX 6000 Ada GPU at batch size~1 with $224\times224$
inputs. FLOPs measure theoretical complexity but do not reflect wall-clock speed;
latency directly captures deployment efficiency under a standardized setting.
VNCT achieves lower latency than VSSD at Micro, Tiny, and Small scales by replacing
directional scan passes with shared global sufficient statistics, and remains
competitive at Base scale.}
\label{tab:latency}
\footnotesize
\setlength{\tabcolsep}{4pt}
\renewcommand{\arraystretch}{0.92}
\begin{tabular}{lc|lc}
\toprule
\multicolumn{2}{c|}{\textbf{Micro}} &
\multicolumn{2}{c}{\textbf{Tiny}} \\
\midrule
Method        & ms/img & Method     & ms/img \\
\midrule
EffNet-B3     & 4.40   & ConvNeXt-T & 7.27 \\
VSSD-M        & 5.80   & Swin-T     & 8.43 \\
\rowcolor{gray!15}
VNCT-Micro    & \textbf{5.25} &
VMamba-T      & 10.09 \\
& &
VSSD-T        & 8.01 \\
\rowcolor{gray!15}
& &
VNCT-Tiny     & \textbf{7.31} \\
\midrule
\multicolumn{2}{c|}{\textbf{Small}} &
\multicolumn{2}{c}{\textbf{Base}} \\
\midrule
Method        & ms/img & Method     & ms/img \\
\midrule
ConvNeXt-S    & 13.03  & ConvNeXt-B & 21.25 \\
VMamba-S      & 18.31  & VMamba-B   & 27.33 \\
VSSD-S        & 13.24  & VSSD-B     & 23.19 \\
\rowcolor{gray!15}
VNCT-Small    & \textbf{12.35} &
VNCT-Base     & \textbf{22.10} \\
\bottomrule
\end{tabular}
\vspace{-2mm}
\end{table}

\subsection{Object Detection and Instance Segmentation on COCO}

\noindent\textbf{Experimental Setup.}
We evaluate on MS COCO~\cite{coco} using Mask R-CNN~\cite{maskrcnn} in
MMDetection~\cite{chen2019mmdetection}, following the protocol of prior
work~\cite{Swin,liu2024vmamba}.

\noindent\textbf{Results and Discussion.}
VNCT consistently improves both box and mask AP over all Vision Mamba architectures
(Table~\ref{tab:coco_vnct}). The largest gains appear at AP$_{75}$, indicating
improved localization precision. Improvements across both detection and segmentation
suggest that global non-causal aggregation and second-order state dynamics yield
stronger spatial representations for dense prediction.

\begin{table}[t]
\centering
\caption{\textbf{Object detection and instance segmentation with Mask R-CNN on
MS COCO~\cite{coco}.}
Box AP (AP$^b$) and mask AP (AP$^m$) under $1\times$ and $3\times$+MS
(multi-scale) training schedules; subscripts 50 and 75 denote IoU thresholds.
Protocol: shorter side 800\,px, longer side $\leq$1333\,px, AdamW with
lr $1\!\times\!10^{-4}$, batch size~16. VNCT-T surpasses all baselines on every
metric at both schedules, with the largest gains at AP$_{75}$.}
\label{tab:coco_vnct}
\footnotesize
\setlength{\tabcolsep}{2pt}
\resizebox{\columnwidth}{!}{
\begin{tabular}{l|cccccc|cccccc}
\toprule
\multicolumn{1}{c|}{\multirow{2}{*}{\textbf{Method}}}
& \multicolumn{6}{c|}{$1\times$}
& \multicolumn{6}{c}{$3\times$ + MS} \\
\cline{2-13}
& AP$^b$ & AP$^b_{50}$ & AP$^b_{75}$ & AP$^m$ & AP$^m_{50}$ & AP$^m_{75}$
& AP$^b$ & AP$^b_{50}$ & AP$^b_{75}$ & AP$^m$ & AP$^m_{50}$ & AP$^m_{75}$ \\
\midrule
Swin-T~\cite{Swin}
  & 42.7 & 65.2 & 46.8 & 39.3 & 62.2 & 42.2
  & 46.0 & 68.1 & 50.3 & 41.6 & 65.1 & 44.9 \\
ConvNeXt-T~\cite{convnext}
  & 44.2 & 66.6 & 48.3 & 40.1 & 63.3 & 42.8
  & 46.2 & 67.9 & 50.8 & 41.7 & 65.0 & 44.9 \\
VMamba-T~\cite{liu2024vmamba}
  & 46.5 & 68.5 & 50.7 & 42.1 & 65.5 & 45.3
  & 48.5 & 69.9 & 52.9 & 43.2 & 66.8 & 46.3 \\
LocalVMamba-T~\cite{huang2024localmamba}
  & 46.7 & 68.7 & 50.8 & 42.2 & 65.7 & 45.5
  & 48.7 & 70.1 & 53.0 & 43.4 & 67.0 & 46.4 \\
\midrule
VSSD-T~\cite{VSSD}
  & 46.9 & 69.4 & 51.4 & 42.6 & 66.4 & 45.9
  & 48.8 & 70.4 & 53.4 & 43.6 & 67.6 & 46.9 \\
\rowcolor{gray!15}
VNCT-T
  & \textbf{47.8} & \textbf{70.3} & \textbf{52.6}
  & \textbf{43.5} & \textbf{67.3} & \textbf{46.8}
  & \textbf{49.7} & \textbf{71.3} & \textbf{54.6}
  & \textbf{44.5} & \textbf{68.6} & \textbf{47.9} \\
\bottomrule
\end{tabular}}
\vspace{-2mm}
\end{table}

\subsection{Semantic Segmentation on ADE20K}

\begin{figure*}[t]
    \centering
    \includegraphics[width=\textwidth]{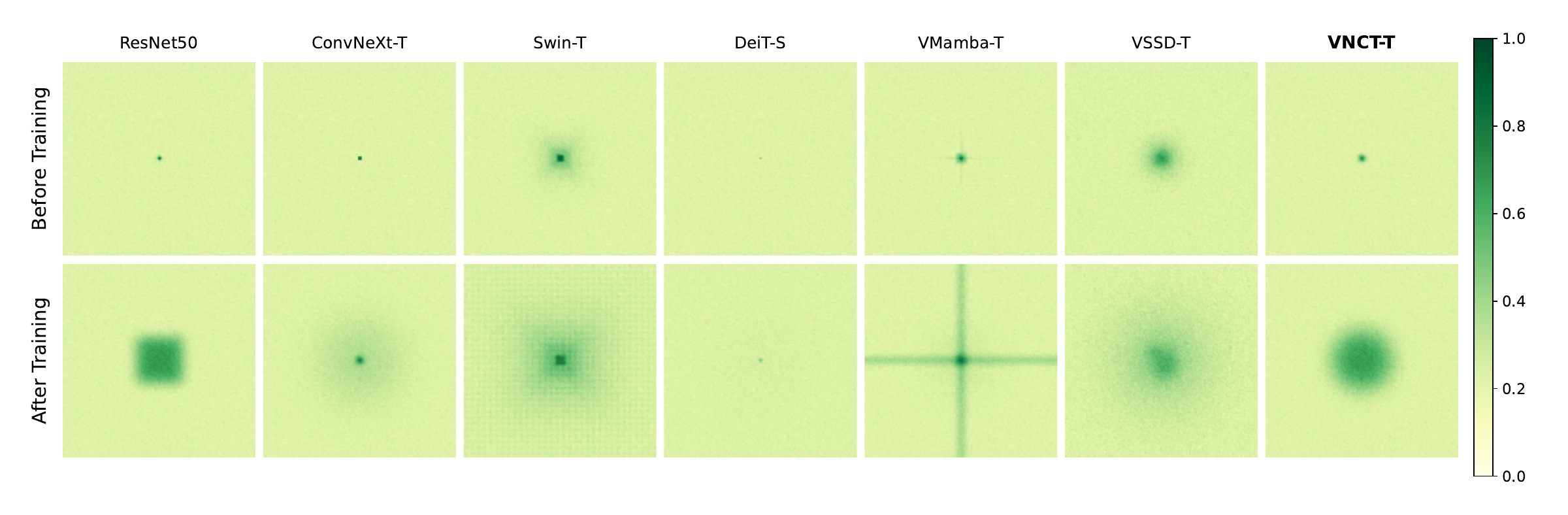}
    \caption{\textbf{Effective Receptive Field (ERF) maps before and after training.}
    Spatial gradient energy of the central pixel with respect to the input image,
    averaged over the ImageNet-1K validation set using the gradient-based method
    of~\cite{luo2016understanding}; values normalized to $[0,1]$. Seven representative
    backbones are compared: CNN-based (ResNet-50, ConvNeXt-T), attention-based
    (Swin-T, DeiT-S), and SSM-based (VMamba-T, VSSD-T, VNCT-T). CNNs produce
    localized receptive fields due to finite convolutional kernels; DeiT-S shows a
    centralized high-response region with weak global activation; VMamba-T exhibits
    prominent horizontal/vertical artifacts from directional scan-based state
    propagation; VSSD-T substantially enlarges coverage but remains irregular and
    asymmetric. VNCT-T uniquely achieves a smooth, symmetric, and uniformly
    distributed receptive field spanning the entire image, reflecting isotropic
    long-range propagation without directional scanning or attention.}
    \label{fig:erf_visualizations}
    \vspace{-2mm}
\end{figure*}

\noindent\textbf{Experimental Setup.}
We evaluate semantic segmentation on ADE20K~\cite{ade20k} with the UPerNet
decoder~\cite{upernet} in MMSegmentation~\cite{mmseg2020}, reporting mIoU
under single-scale (SS) and multi-scale (MS) evaluation.

\noindent\textbf{Results and Discussion.}
VNCT achieves the strongest segmentation performance among comparable state-space
backbones at all model scales (Table~\ref{tab:ade20k_vnct}). Consistent improvements
over other SSM-based architectures indicate that the gains are not solely attributable
to removing directional scanning; the results further suggest that second-order
non-causal dynamics are particularly beneficial for dense scene understanding and
fine-grained spatial modeling.

\begin{table}[t]
\centering
\caption{\textbf{Semantic segmentation on ADE20K~\cite{ade20k} with
UPerNet~\cite{upernet}.}
Mean Intersection-over-Union (mIoU, \%) under single-scale (SS) and multi-scale
(MS) evaluation. All backbones initialized from ImageNet-1K pretrained checkpoints
and trained for 160K iterations under identical optimization settings in
MMSegmentation~\cite{mmseg2020}. Parameter and FLOPs counts include the UPerNet
decoder. VNCT achieves the highest mIoU at all three model scales (Micro, Tiny,
Small), consistently outperforming both scan-based and first-order non-causal
SSM backbones.}
\label{tab:ade20k_vnct}
\footnotesize
\setlength{\tabcolsep}{3pt}
\renewcommand{\arraystretch}{0.8}
\begin{tabular}{l|cc|cc}
\toprule
\textbf{Method} & mIoU (SS) & mIoU (MS) & \#Param. & FLOPs \\
\midrule
EffVMamba-S~\cite{pei2024efficientvmamba}  & 41.5 & 42.1 & 29M & 505G \\
MSVMamba-M~\cite{shi2024multi}             & 45.1 & 45.4 & 42M & 875G \\
VSSD-M~\cite{VSSD}                         & 45.6 & 46.0 & 42M & 893G \\
VNCT-M                                     & \textbf{46.5} & \textbf{47.0} & 43M & 900G \\
\midrule
Swin-T~\cite{Swin}                         & 44.4 & 45.8 & 60M &  945G \\
ConvNeXt-T~\cite{convnext}                 & 46.0 & 46.7 & 60M &  939G \\
VMamba-T~\cite{liu2024vmamba}              & 47.3 & 48.3 & 55M &  964G \\
LocalVMamba-T~\cite{huang2024localmamba}   & 47.9 & 49.1 & 57M &  970G \\
EffVMamba-B~\cite{pei2024efficientvmamba}  & 46.5 & 47.3 & 65M &  930G \\
MSVMamba-T~\cite{shi2024multi}             & 47.6 & 48.5 & 65M &  942G \\
VSSD-T~\cite{VSSD}                         & 47.9 & 48.7 & 53M &  941G \\
VNCT-T                                     & \textbf{48.8} & \textbf{49.7} & 54M & 950G \\
\midrule
Swin-S~\cite{Swin}                         & 47.6 & 49.5 & 81M & 1038G \\
VMamba-S~\cite{liu2024vmamba}              & 49.5 & 50.5 & 76M & 1081G \\
VSSD-S~\cite{VSSD}                         & 49.2 & 50.1 & 69M & 1002G \\
VNCT-S                                     & \textbf{50.1} & \textbf{51.2} & 44M & 1015G \\
\bottomrule
\end{tabular}
\vspace{-2mm}
\end{table}

\subsection{Effective Receptive Field Visualization}
\label{subsec:erf_analysis}

\noindent\textbf{Experimental Setup.}
We visualize the Effective Receptive Field (ERF) following~\cite{luo2016understanding},
computing the gradient magnitude of the central pixel with respect to the input across
the ImageNet-1K validation set, both before and after training.

\noindent\textbf{Results and Discussion.}
The ERF maps in Fig.~\ref{fig:erf_visualizations} reveal the distinct spatial inductive
biases of each architecture. VNCT-T is the only model to produce a smooth, symmetric,
and globally uniform receptive field, indicating that the proposed second-order
exponential-trapezoidal dynamics promote balanced long-range information propagation
without directional scans or global attention.

\subsection{Orientation Robustness on ImageNet-1K}

\noindent\textbf{Experimental Setup.}
We assess model robustness under common geometric transformations of the ImageNet-1K
validation set without additional training or test-time adaptation.

\noindent\textbf{Results and Discussion.}
VNCT exhibits the strongest robustness across all transformations and the smallest
average accuracy drop (Table~\ref{tab:orientation_robustness}). The reduced degradation
under flips and rotations confirms that learned representations are less
orientation-sensitive; comparison with VSSD further indicates that improved robustness
stems not only from removing directional scans but also from the proposed second-order
non-causal dynamics.

\begin{table}[t]
\centering
\caption{\textbf{Orientation robustness on the ImageNet-1K validation set.}
Top-1 accuracy (\%) evaluated without retraining or test-time adaptation under four
geometric transformations: Horizontal Flip (HF), Vertical Flip (VF), $90^{\circ}$
Rotation (R90), and $180^{\circ}$ Rotation (R180). Avg.\ Drop: mean accuracy
decrease across all four variants relative to the original images; lower is better
and indicates reduced orientation sensitivity. VNCT-T achieves the highest accuracy
on every variant and the smallest average drop (0.3\%), compared to VSSD-T (0.9\%)
and VMamba-T (1.6\%), demonstrating more isotropic spatial representations
attributable to both the removal of directional scanning and the second-order
non-causal dynamics.}
\label{tab:orientation_robustness}
\footnotesize
\setlength{\tabcolsep}{2pt}
\renewcommand{\arraystretch}{0.95}
\resizebox{\columnwidth}{!}{
\begin{tabular}{lccccccc}
\toprule
Method & Orig. & HF & VF & R90 & R180 & Avg.\ Drop \\
\midrule
VMamba-T~\cite{liu2024vmamba}
  & 82.5 & 82.1 & 80.8 & 79.6 & 81.0 & 1.6 \\
VSSD-T~\cite{VSSD}
  & 83.7 & 83.4 & 82.7 & 81.9 & 82.8 & 0.9 \\
\rowcolor{gray!15}
VNCT-T
  & \textbf{84.2} & \textbf{84.1} & \textbf{83.9}
  & \textbf{83.5} & \textbf{83.9} & \textbf{0.3} \\
\bottomrule
\end{tabular}}
\vspace{-2mm}
\end{table}

\begin{figure*}[t]
    \centering
    \includegraphics[width=\textwidth]{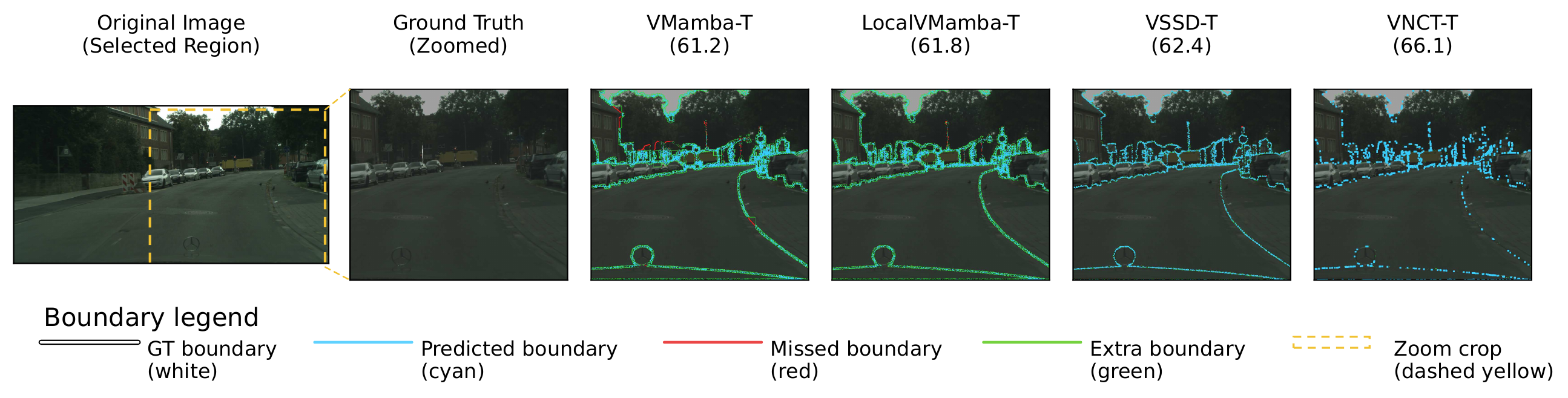}
    \caption{\textbf{Qualitative boundary predictions on Cityscapes.}
    A road-scene crop comparing boundary outputs for VMamba-T, LocalVMamba-T,
    VSSD-T, and VNCT-T against the ground-truth contour. Color code: cyan =
    correctly predicted boundaries; red = missed boundaries; green = spurious
    boundaries. Yellow dashed box indicates the zoomed region. VNCT-T produces
    the most complete and geometrically accurate contour alignment, consistent
    with its superior BIoU scores in Table~\ref{tab:boundary_results}.}
    \label{fig:boundary_iou_results}
    \vspace{-4mm}
\end{figure*}

\subsection{Boundary-Sensitivity Evaluation}

\noindent\textbf{Experimental Setup.}
Region-based metrics (AP, mIoU) primarily measure area overlap and may not fully
capture contour quality. Following Cheng~\emph{et al.}~\cite{cheng2021boundary},
we additionally report Boundary IoU (BIoU) on ADE20K and Cityscapes to evaluate
prediction quality near object contours.

\noindent\textbf{Results and Discussion.}
VNCT-T achieves highest BIoU on both benchmarks (Table~\ref{tab:boundary_results}),
with gains of $+2.3$ and $+3.7$ BIoU over VSSD-T on ADE20K and Cityscapes,
respectively. These gains complement the mIoU improvements and confirm that
second-order non-causal dynamics enhance fine-grained contour localization alongside
semantic region prediction. Qualitative results in Fig.~\ref{fig:boundary_iou_results} illustrate the sharper and more complete boundary alignment achieved by VNCT.

\begin{table}[!ht]
\centering
\caption{\textbf{Boundary IoU (BIoU) on ADE20K and Cityscapes.}
BIoU~\cite{cheng2021boundary} evaluates segmentation quality near object contours,
complementing region-based metrics (mIoU, AP) that primarily measure area overlap.
Higher is better. VNCT-T outperforms all Mamba-based baselines on both datasets,
with gains of $+2.3$ and $+3.7$ BIoU over VSSD-T on ADE20K and Cityscapes,
respectively.}
\label{tab:boundary_results}
\footnotesize
\begin{tabular}{lcc}
\toprule
Method & ADE20K & Cityscapes \\
\midrule
VMamba-T~\cite{liu2024vmamba}           & 48.3 & 61.2 \\
LocalVMamba-T~\cite{huang2024localmamba}& 48.7 & 61.8 \\
VSSD-T~\cite{VSSD}                      & 49.1 & 62.4 \\
VNCT-T                                  & \textbf{51.4} & \textbf{66.1} \\
\bottomrule
\end{tabular}
\vspace{-2mm}
\end{table}

\subsection{Ablation Studies}

Unless otherwise specified, all ablations use the VNCT-T backbone. Table~\ref{tab:ablation} summarizes the principal design ablations. Each proposed component contributes to the final performance, with the largest gains arising from non-causal aggregation and the proposed second-order state dynamics. Additional ablations and details are in Supplement~\ref{sec:supp_ablation}.

\begin{table}[t]
\centering
\caption{\textbf{Principal ablations on VNCT-T.} \emph{Left:} core components, evaluated across all tasks---Top-1 (\%, ImageNet-1K), box AP (COCO~$1\times$), and mIoU/BIoU (\%, ADE20K). \emph{Right:} positional encoding and state dynamics, evaluated on Top-1 and mIoU. Best per group in \textbf{bold}.}
\label{tab:ablation}
\scriptsize
\setlength{\tabcolsep}{2pt}
\renewcommand{\arraystretch}{0.95}
\begin{tabular}[c]{@{}lcccc@{}}
\toprule
\multicolumn{5}{c}{\textit{Core Components}} \\
\midrule
Variant & Top-1 & AP & mIoU & BIoU \\
\midrule
Scan & 83.5 & 46.4 & 47.6 & 48.9 \\
SISO & 83.8 & 47.0 & 48.0 & 49.5 \\
\rowcolor{gray!15}
VNCT & \textbf{84.2} & \textbf{47.8} & \textbf{48.8} & \textbf{50.8} \\
\bottomrule
\end{tabular}
\hspace{0.6em}
\begin{tabular}[c]{@{}lcc@{}}
\toprule
\multicolumn{3}{c}{\textit{Positional Encoding}} \\
\midrule
Variant & Top-1 & mIoU \\
\midrule
None & 82.9 & 46.8 \\
2D Sin. & 83.7 & 48.0 \\
RoPE & 83.8 & 48.1 \\
\rowcolor{gray!15}
2D Sin.+RoPE & \textbf{84.2} & \textbf{48.8} \\
\midrule
\multicolumn{3}{c}{\textit{State Dynamics}} \\
\midrule
VSSD-T (1st) & 83.7 & 47.9 \\
\rowcolor{gray!15}
VNCT-T (2nd) & \textbf{84.2} & \textbf{48.8} \\
\bottomrule
\end{tabular}
\vspace{-2mm}
\end{table}

\section{Conclusion}
We revisit assumption that vision state-space models require the directional scanning inherited from sequence modeling, and show they do not: pairing scan-free global aggregation with second-order trapezoidal dynamics, VNCT improves recognition, dense prediction, boundary quality at efficient inference. This motivates non-causal, higher-order state-space operators as promising direction for vision backbones.
\raggedbottom

{
    \small
    \bibliographystyle{ieeenat_fullname}
    \bibliography{main}
}

\clearpage
\appendix

\section{Supplementary Material}


\subsection{Additional Related Work}
\label{subsec:supp_related_work}

\subsubsection{Vision State-Space Backbones and the Scanning Problem}

State-space models (SSMs) provide a linear-complexity alternative to self-attention for long-range sequence modeling, beginning with structured SSMs such as S4 and later selective models such as Mamba and Mamba-2/SSD~\cite{gu2022efficient,gu2023mamba,dao2024transformers}. Their extension to vision has produced a rapidly growing family of visual backbones. S4ND first explored multidimensional state-space modeling for visual signals~\cite{nguyen2022s4nd}, while ViM, VMamba, LocalMamba, EfficientVMamba, PlainMamba, and Multi-Scale VMamba established that Mamba-style token mixers can compete with CNNs and Vision Transformers on classification and dense prediction~\cite{zhu2024ViM,liu2024vmamba,huang2024localmamba,pei2024efficientvmamba,plainmamba,shi2024multi}. However, most of these models inherit the causal one-dimensional recurrence of sequence models: images are flattened into token sequences, processed along horizontal, vertical, local, or multi-directional scan paths, and then reassembled into two-dimensional feature maps. These scan routes partially compensate for missing future-token context, but they also impose an artificial spatial order that is not intrinsic to images.

A second line of work improves the scan itself rather than removing it. DAMamba learns input-adaptive scan orders and regions, DefMamba introduces deformable scan paths, QuadMamba uses quadtree-based selective scanning, and GroupMamba partitions channels into grouped directional scans with post-hoc channel affinity modulation~\cite{li2025damamba,liu2025defmamba,xie2024quadmamba,shaker2025groupmamba}. SF-Mamba further revisits the efficiency of visual Mamba by retaining a unidirectional causal scan while routing future-to-past information through swapped auxiliary tokens and improving GPU utilization through batch folding~\cite{yoshimura2026sfmamba}. These designs demonstrate that scan order is a central bottleneck in vision SSMs, yet they still treat visual modeling as an ordered traversal problem. In contrast, VNCT removes the traversal itself: NC-M3 aggregates over all image tokens through a scan-order-independent non-causal operator, so there is no learned, handcrafted, swapped, or deformable scan path to optimize.

\subsubsection{Scan-Free and Non-Causal Vision SSMs}

The closest line of work to VNCT explores whether the causal mask in visual SSMs can be removed. VSSD introduced a non-causal form of State Space Duality for vision, replacing token-wise causal hidden states with a shared global state and showing that explicit scan routes are not necessary for strong visual performance~\cite{VSSD}. However, VSSD is built from first-order Mamba-2/SSD dynamics and uses a SISO-style state parameterization, whereas VNCT lifts Mamba-3's second-order exponential-trapezoidal dynamics into a non-causal visual operator and further introduces low-rank MIMO state mixing.

Several recent methods reduce or bypass scanning through other mechanisms. Spatial-Mamba keeps a single causal scan and restores two-dimensional structure through local state fusion with dilated depthwise convolutions~\cite{xiao2025spatialmamba}; Mamba2D derives a native two-dimensional recurrence but still evaluates it through a causal wavefront scan~\cite{baty2024mamba2d}; PRISMamba averages tokens within concentric rings but retains a residual radial scan across rings~\cite{hsieh2026partialring}. Other scan-free designs move away from state-space recurrence: SEMA combines local window attention with a global arithmetic average of value tokens~\cite{tran2025sema}, while HAMSA replaces the SSM recurrence with FFT-domain spectral mixing~\cite{patro2026hamsa}. DensePercept-NCSSD extends non-causal SSD ideas to dense perception tasks such as optical flow and disparity estimation~\cite{anand2025densepercept}. These works support the broader conclusion that images need not be processed through directional recurrence, but differ in what replaces the scan. VNCT instead preserves an SSM interpretation by parameterizing global aggregation through second-order state-space coefficients.

\subsubsection{SSM Operator Expressivity and Second-Order Dynamics}

The connection between SSMs and attention has become increasingly explicit. Mamba-2 formulated selective SSMs through State Space Duality as structured masked matrix mixers~\cite{dao2024transformers}, while Mamba-3 introduced exponential-trapezoidal discretization, complex-valued state updates, and a MIMO formulation~\cite{lahoti2026mamba}. VNCT adopts this second-order viewpoint but reformulates the recurrence as a non-causal aggregation operator for images.

Recent theoretical and empirical studies suggest that richer transition operators improve SSM expressivity. Merrill et al. analyze limitations of diagonal first-order SSMs, while negative-eigenvalue formulations expand state-tracking capability in linear recurrent models~\cite{merrill2024illusion,grazzi2025negeigen}. Gated DeltaNet further enriches Mamba-style operators through delta-rule memory updates~\cite{yang2025gateddelta}. VNCT is complementary to these directions, exploring second-order trapezoidal dynamics within a scan-free non-causal visual formulation.

\subsubsection{Efficient Global Mixing, Linear Attention, and Hybrid Designs}

Because VNCT factorizes global source statistics followed by per-token readout, it is algebraically related to linear attention~\cite{katharopoulos2020transformers,choromanski2020rethinking}, consistent with recent analyses relating Mamba-style models to kernelized attention~\cite{han2024demystify}. Unlike conventional linear attention, however, VNCT parameterizes its kernel using data-dependent SSM discretization coefficients, including the trapezoidal $\alpha,\beta,\gamma$ parameters.

Efficient global mixers such as FLatten, Gated Linear Attention, Vision-RWKV, and RALA improve linear token mixing through focused kernels, gating, recurrent mixing, or rank augmentation~\cite{han2023flatten,yang2024gla,duan2025visionrwkv,fan2025rala}. VNCT instead introduces cross-channel interaction through a low-rank MIMO state-space formulation. Hybrid architectures including MambaVision, SegMAN, and A2Mamba combine SSMs with attention modules~\cite{hatamizadeh2025mambavision,fu2025segman,lou2025a2mamba}, whereas MambaOut questions whether SSM recurrence itself is necessary for visual recognition~\cite{yu2024mambaout}. VNCT takes a different perspective: the limitation lies in directional scanning rather than state-space modeling, motivating a second-order non-causal formulation with attention restricted to the final low-resolution stage.

\subsection{Notation}
\label{sec:supp_notation}


Let $\mathbf{X} = [\mathbf{x}_1, \ldots, \mathbf{x}_L]^\top \in \mathbb{R}^{L
\times C}$ denote a sequence of $L$ image tokens with channel dimension $C$,
obtained by flattening a spatial feature map of resolution $H \times W$, so that
$L = HW$. We index tokens with $i$ (target) and $j$ (source), and channels with
$c$. Vectors are bold lowercase; matrices and tensors are bold uppercase.
Throughout, we treat each channel independently when deriving scalar-valued
aggregation expressions, then lift to the full channel dimension by broadcasting.

The following additional symbols are used consistently: $N$ denotes the SSM state
dimension, $H$ the number of SSM heads, $R$ the MIMO rank,
$d_{\mathrm{head}} = d_{\mathrm{inner}} / H$ the per-head feature dimension with
$d_{\mathrm{inner}} = 2C$ by default, and $S$ the number of RoPE angle
dimensions.

\subsection{Causal Bias Analysis: Why Lower-Triangular Structure Is
Problematic for Vision}
\label{sec:supp_causal}


To understand the motivation concretely, consider unrolling Eq.~\eqref{eq:mamba3} over $L$
steps. For a single channel $c$, defining $\tilde{x}_{j,c} = b_{j,c}\,x_{j,c}$,
the causal state sequence is described by a lower-triangular linear map
\begin{equation}
    \mathbf{h}_c = \mathbf{M}_c\,\tilde{\mathbf{x}}_c,
    \label{eq:transition_matrix}
\end{equation}
where $\mathbf{h}_c, \tilde{\mathbf{x}}_c \in \mathbb{R}^L$ collect per-position
states and weighted observations, and $\mathbf{M}_c \in \mathbb{R}^{L \times L}$
is the transition matrix with $M_{ij,c} = 0$ whenever $j > i$. For a
two-dimensional image flattened into a one-dimensional sequence, this triangular
structure has no relationship to spatial proximity: two adjacent pixels separated
by the scan fold contribute asymmetrically to each other's state, while two
distant tokens on the same scan-row interact more strongly than geometrically
closer tokens on different rows. Multi-directional approaches~\cite{liu2024vmamba}
mitigate this by averaging over multiple lower-triangular maps, but each
individual map remains lower-triangular and their average does not achieve
permutation invariance. The non-causal lift replaces $\mathbf{M}_c$ with a dense,
symmetric interaction matrix derived from the same parameters $\boldsymbol{\alpha}$,
$\boldsymbol{\beta}$, $\boldsymbol{\gamma}$.

\subsection{2D RoPE: Full Derivation and Implementation}
\label{sec:supp_rope}


For a token at grid position $(r_i, c_i)$, the phase vector is
\begin{equation}
\begin{aligned}
    \boldsymbol{\varphi}_i =
    \Bigl[
    &r_i\,f_{h,1},\;\ldots,\;r_i\,f_{h,S/2},\; \\
    &c_i\,f_{w,1},\;\ldots,\;c_i\,f_{w,S/2}
    \Bigr] \in \mathbb{R}^S,
\end{aligned}
    \label{eq:rope_phase}
\end{equation}
where $f_{h,k} = f_{w,k} = 10000^{-2(k-1)/S}$ are the standard sinusoidal
frequency terms. The rotation is applied to adjacent pairs in the last dimension:
\begin{equation}
    \tilde{B}_{i,r,s}
    = B_{i,r,s}\cos\varphi_{i,s} - B_{i,r,s+S/2}\sin\varphi_{i,s},
    \label{eq:rope_apply}
\end{equation}
and symmetrically for $\mathbf{C}$, where $r$ indexes MIMO rank and $s$ indexes
the phase dimension. The inner product
$\tilde{\mathbf{B}}_{j,r}^{\top}\tilde{\mathbf{C}}_{i,r}$ that appears
implicitly in the aggregation einsum becomes
\begin{equation}
\begin{aligned}
\tilde{\mathbf{B}}_{j,r}^{\top}\tilde{\mathbf{C}}_{i,r}
=
\sum_{s=1}^{S/2}
\Bigl[
&B_{j,r,s}C_{i,r,s}\cos(\varphi_{j,s}-\varphi_{i,s})
\\
&+ B_{j,r,s}C_{i,r,s+S/2}\sin(\varphi_{j,s}-\varphi_{i,s}) \\
&+ \cdots
\Bigr],
\end{aligned}
\label{eq:rope_inner}
\end{equation}
which depends on phase differences $\varphi_{j,s}-\varphi_{i,s}$, encoding
relative spatial displacement. Crucially, after rotation $\tilde{\mathbf{B}}$
depends only on source index $j$ and $\tilde{\mathbf{C}}$ only on target index
$i$, so the global statistics $\mathbf{S}_c$ and $\mathbf{Z}_c$ can still be
accumulated in a single pass. In implementation, phases are computed from the
actual $(H,W)$ grid dimensions at each stage. The remaining dimensions of
$\mathbf{B}$ and $\mathbf{C}$ beyond $S$ phase pairs are left unrotated.

\subsection{Scan Invariance vs.\ Direction Invariance}
\label{sec:supp_scan}

\begin{figure}[t]
    \centering
    \includegraphics[width=\columnwidth]{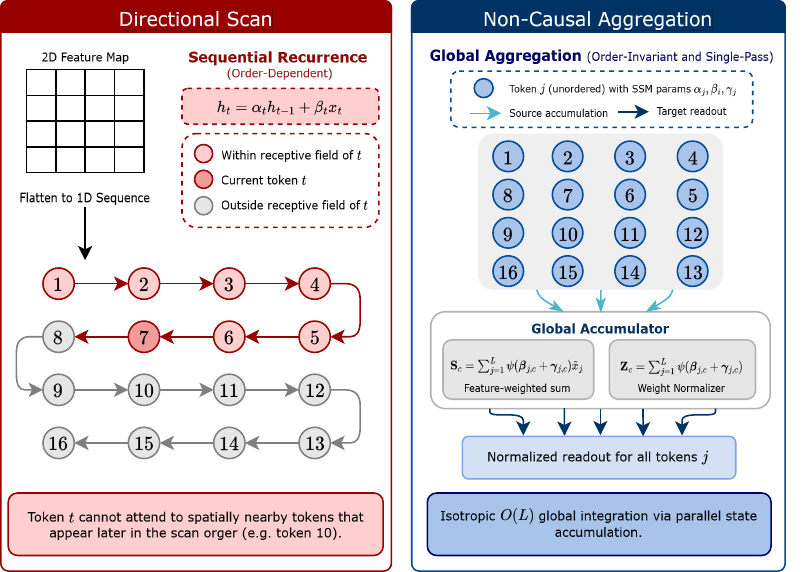}
    \caption{\textbf{Directional scanning versus non-causal aggregation.} 
    Existing vision SSMs impose one or more fixed scan orders over the token 
    grid (e.g.\ horizontal, vertical, or multi-directional), introducing an 
    artificial sequential ordering that is not intrinsic to images. NC-M3 
    instead aggregates all tokens into a single shared global state that every 
    token reads back, eliminating scan order entirely while preserving 
    linear-time complexity.}
    \label{fig:scan_overview}
\end{figure}

Two related but distinct properties appear in the scan-free vision SSM
literature. \textit{Direction invariance} means the output representation is
insensitive to the orientation of a scan: rotating or reflecting the scan pattern
leaves the representation unchanged. This is approximately achieved by averaging
over multiple fixed scan directions, as in VMamba~\cite{liu2024vmamba} (four
directions) or PlainMamba~\cite{plainmamba} (continuous 2D scanning). Direction
invariance does not imply invariance to arbitrary token permutations.
\textit{Scan invariance} is the strictly stronger property that the output is
invariant to any permutation of the input token sequence. NC-M3 achieves scan
invariance by construction: the accumulation in Eqs.~\eqref{eq:G}--\eqref{eq:readout} is a commutative
sum and the readout draws from the same global $\mathbf{G}$ for every target
token. Direction-invariant methods may exhibit residual anisotropy in effective
receptive field (ERF) patterns~\cite{VSSD}, whereas scan-invariant methods have
isotropic ERFs by construction, which is advantageous for boundary detection and
small-object segmentation. Spatial structure is re-introduced into NC-M3 through
2D RoPE (relative position) and Pos2D (absolute position), consistent with
VSSD~\cite{VSSD} and PlainMamba~\cite{plainmamba}.
Figure~\ref{fig:scan_overview} contrasts the two regimes: directional vision
SSMs traverse the token grid along one or more fixed scan orders, whereas
NC-M3 replaces every such traversal with a single permutation-invariant
aggregation into a shared global state that all tokens read back.

\subsection{MIMO: Extended Analysis}
\label{sec:supp_mimo}


The parameter and computational cost of MIMO is $\mathcal{O}(HRd_{\mathrm{head}})
= \mathcal{O}(CR)$ for the mixing tensor $\mathbf{U}$, compared to
$\mathcal{O}(C^2)$ for a full cross-channel linear map. For typical values
($R=4$, $C=96$--$768$), this represents a parameter reduction of $24\times$ to
$192\times$ relative to a full mixing layer. Without MIMO ($R=1$), the
aggregation reduces to the standard SISO form and $\mathbf{U}$ degenerates to a
per-head identity.

\subsection{Memory-Efficient Implementation}
\label{sec:supp_memory}


To avoid materializing the full $(B, L, H, d_{\mathrm{head}}, R, N)$
intermediate tensor, both Eqs.~\eqref{eq:G} and~\eqref{eq:readout} are evaluated in chunks of $K$
tokens along the spatial dimension, requiring $\mathcal{O}(K\cdot H\cdot
d_{\mathrm{head}}\cdot R\cdot N)$ peak memory per chunk. The chunk size $K$ is
set adaptively based on model width: $K=256$ for micro and tiny variants,
$K=128$ for small, and $K=32$ for base, calibrated to keep peak per-chunk memory below 150\,MB on A100-40\,GB at the respective batch sizes.

The step-size bias $b_{\Delta,h}$ in Eq.~\eqref{eq:dt} is initialized following the
Mamba-2 strategy~\cite{dao2024transformers} to encourage moderate initial step
sizes.

\subsection{Hierarchical Design Rationale}
\label{sec:supp_hierarchy}


The motivation for hierarchical processing is twofold. First, dense prediction
tasks require features at multiple scales: fine-grained spatial detail at high resolution for precise boundary localization and abstract semantic content at low
resolution for region classification. Second, hierarchical processing enables
NC-M3 to work at progressively smaller sequence lengths in deeper stages; at
lower resolutions, each token represents a larger spatial region and aggregation
across all tokens captures broader context with higher signal-to-noise ratio.

The LPU serves a secondary purpose beyond injecting local inductive bias:
by providing each token with awareness of its immediate neighborhood prior to RoPE-based global aggregation, it reduces the burden on the positional encoding to capture short-range dependencies.
\subsection{VNCT Backbone Configurations}
\label{sec:supp_backbone}

Table~\ref{tab:vnct_backbone_all} summarizes the backbone configurations of all VNCT variants. All models share the same architecture shown in Fig.~\ref{fig:vnct_backbone}: Stages 1--3 employ VNCT (NC-M3) blocks, while Stage~4 uses Hybrid-Attention. The variants differ only in channel dimensions, stage depths, and stochastic depth rate. Unless otherwise specified, all NC-M3 layers use a state dimension of $d_{\mathrm{state}}=64$.

\begin{table*}[t]
\centering
\caption{\textbf{VNCT backbone configurations.} Stage resolutions are fixed at $\{H/4,H/8,H/16,H/32\}$ for all variants.}
\label{tab:vnct_backbone_all}
\small
\setlength{\tabcolsep}{8pt}
\renewcommand{\arraystretch}{1.1}
\begin{tabular}{lccc}
\toprule
\textbf{Variant} &
\textbf{Stage Channels} &
\textbf{Stage Depths} &
\textbf{Drop Path} \\
\midrule
VNCT-Micro &
$\{64,128,256,512\}$ &
$\{2,2,6,2\}$ &
0.1 \\
VNCT-Tiny &
$\{96,192,384,768\}$ &
$\{2,2,9,2\}$ &
0.2 \\
VNCT-Small &
$\{96,192,384,768\}$ &
$\{3,3,18,3\}$ &
0.3 \\
VNCT-Base &
$\{128,256,512,1024\}$ &
$\{3,3,27,3\}$ &
0.5 \\
\bottomrule
\end{tabular}
\end{table*}
\subsection{Training Details}

Unless otherwise specified, ImageNet-1K models are trained for 300 epochs using AdamW with an initial learning rate of $1\times10^{-3}$, weight decay of 0.05, a cosine learning-rate schedule, and a 20-epoch linear warm-up. Models are trained at $224\times224$ resolution with a per-GPU batch size of 128 (64 for VNCT-Base using gradient accumulation). We employ RandAugment, Mixup, CutMix, Random Erasing, and label smoothing. Exponential Moving Average (EMA) weights with a decay factor of 0.9999 are maintained throughout training, and the checkpoint with the highest validation Top-1 accuracy is used for evaluation.

For downstream object detection and instance segmentation, we adopt the official VSSD configuration using Mask R-CNN in MMDetection under the standard $1\times$ and $3\times$ schedules. Semantic segmentation follows the official UPerNet configuration in MMSegmentation. All downstream experiments initialize the backbone from the corresponding ImageNet-1K pretrained checkpoint while keeping all optimization settings identical to the VSSD baseline.

\subsection{Extended Ablation Studies}
\label{sec:supp_ablation}

Unless otherwise specified, all ablation studies are conducted using the VNCT-T backbone. This appendix provides detailed analyses of the architectural choices summarized in the main paper, including non-causal aggregation, second-order dynamics, the proposed low-rank MIMO formulation, positional encoding, and comparisons with first-order non-causal state-space models.

\subsection{Effect of Non-Causal Aggregation}

To evaluate the role of global non-causal aggregation, we compare VNCT against a causal counterpart in which non-causal aggregation is replaced with directional scan-based recurrence while keeping all other components unchanged. As shown in Table~\ref{tab:ablation_noncausal}, replacing directional recurrence with non-causal aggregation consistently improves performance across all evaluated tasks. The largest gains are observed on dense prediction benchmarks, particularly Boundary IoU, demonstrating that isotropic global interactions provide richer spatial context than directional scan paths.

\begin{table}[t]
\centering
\caption{\textbf{Effect of non-causal aggregation.}}
\label{tab:ablation_noncausal}
\footnotesize
\begin{tabular}{lcc}
\toprule
Metric & Scan-based & Non-causal \\
\midrule
Top-1 (\%) & 83.5 & \textbf{84.2} \\
COCO AP & 46.4 & \textbf{47.8} \\
ADE mIoU & 47.6 & \textbf{48.8} \\
Boundary IoU & 48.9 & \textbf{50.8} \\
\bottomrule
\end{tabular}
\end{table}

\subsection{Effect of Second-Order Dynamics}

To isolate the contribution of the proposed exponential-trapezoidal formulation, we replace the second-order update with a first-order Euler approximation while keeping the remaining architecture unchanged. Results in Table~\ref{tab:ablation_secondorder} show consistent improvements across both classification and dense prediction tasks. In particular, the larger gain in Boundary IoU suggests that second-order dynamics better preserve fine spatial structures and object boundaries by capturing richer state evolution.

\begin{table}[t]
\centering
\caption{\textbf{Effect of second-order dynamics.}}
\label{tab:ablation_secondorder}
\footnotesize
\begin{tabular}{lcc}
\toprule
Metric & First-order & Second-order \\
\midrule
Top-1 (\%) & 83.6 & \textbf{84.2} \\
ADE mIoU & 47.7 & \textbf{48.8} \\
Boundary IoU & 49.2 & \textbf{50.8} \\
\bottomrule
\end{tabular}
\end{table}

\subsection{Effect of the Low-Rank MIMO Formulation}

To assess the effectiveness of the proposed low-rank MIMO parameterization, we replace it with a single-input single-output (SISO) formulation while preserving the same model capacity wherever possible. As reported in Table~\ref{tab:ablation_mimo}, jointly modeling spatial and channel-wise interactions consistently improves performance across all benchmarks. 

\begin{table}[t]
\centering
\caption{\textbf{Effect of the low-rank MIMO formulation.}}
\label{tab:ablation_mimo}
\footnotesize
\begin{tabular}{lcc}
\toprule
Metric & SISO & MIMO \\
\midrule
Top-1 (\%) & 83.8 & \textbf{84.2} \\
COCO AP & 47.0 & \textbf{47.8} \\
ADE mIoU & 48.0 & \textbf{48.8} \\
Boundary IoU & 49.5 & \textbf{50.8} \\
\bottomrule
\end{tabular}
\end{table}

\subsection{Effect of Positional Encoding}

Since non-causal aggregation does not impose an inherent ordering over tokens, positional information plays an important role in modeling spatial relationships. We compare several positional encoding strategies, including no positional encoding, 2D sinusoidal encoding, RoPE, and their combination. As shown in Table~\ref{tab:ablation_pos}, combining 2D sinusoidal encoding with RoPE yields the strongest results, suggesting that the two encoding schemes provide complementary spatial cues.

\begin{table}[ht]
\centering
\caption{\textbf{Effect of positional encoding.}}
\label{tab:ablation_pos}
\footnotesize
\begin{tabular}{lcc}
\toprule
Variant & Top-1 & ADE mIoU \\
\midrule
None & 82.9 & 46.8 \\
2D Sinusoidal & 83.7 & 48.0 \\
RoPE & 83.8 & 48.1 \\
2D Sinusoidal + RoPE & \textbf{84.2} & \textbf{48.8} \\
\bottomrule
\end{tabular}
\end{table}

\subsection{Comparison with First-Order Non-Causal State-Space Models}
\begin{table}[h]
\centering
\caption{\textbf{First-order versus second-order non-causal state-space models.}}
\label{tab:ablation_vssd}
\footnotesize
\begin{tabular}{lcc}
\toprule
Metric & VSSD-T & VNCT-T \\
\midrule
Top-1 (\%) & 83.7 & \textbf{84.2} \\
COCO AP & 46.9 & \textbf{47.8} \\
ADE mIoU & 47.9 & \textbf{48.8} \\
Boundary IoU & 49.1 & \textbf{50.8} \\
\bottomrule
\end{tabular}
\end{table}
We directly compare VNCT-T with the first-order non-causal VSSD-T baseline under identical training and evaluation settings. As summarized in Table~\ref{tab:ablation_vssd}, the proposed second-order formulation consistently outperforms its first-order counterpart across all benchmarks. The largest improvements are observed on dense prediction tasks, with gains of $+0.9$ AP, $+0.9$ mIoU, and $+3.7$ Boundary IoU, indicating that second-order state dynamics provide a more expressive representation for capturing complex spatial dependencies.

\end{document}